\PassOptionsToPackage{hyphens}{url} 
\documentclass[conference]{IEEEtran}
\usepackage{cite}
\usepackage{url}       
\usepackage[hyphens]{url}
\usepackage{hyperref}
\usepackage{booktabs} 
\usepackage{amsmath,amssymb,amsfonts}
\usepackage{algorithmic}
\usepackage{graphicx}
\usepackage{textcomp}
\begin{document}
\title{Analysis of Invasive Breast Cancer in Mammograms Using YOLO, Explainability, and Domain Adaptation}
\author{Jayan Adhikari, Prativa Joshi, and Sushish Baral
}

\maketitle

\begin{abstract}
Deep learning models for breast cancer detection from mammographic images have significant reliability problems when presented with Out-of-Domain (OOD) inputs such as other imaging modalities (CT, MRI, X-ray) or equipment variations, leading to unreliable detection and misdiagnosis. The current research mitigates the fundamental OOD issue through a comprehensive approach integrating ResNet50-based OOD filtering with YOLO architectures (YOLOv8, YOLOv11, YOLOv12) for accurate detection of breast cancer.
Our strategy establishes an in-domain gallery via cosine similarity to rigidly reject non-mammographic inputs prior to processing, ensuring that only domain-associated images supply the detection pipeline. The OOD detection component achieves 99.77\% general accuracy with immaculate 100\% accuracy on OOD test sets, effectively eliminating irrelevant imaging modalities. ResNet50 was selected as the optimum backbone after 12 CNN architecture searches. The joint framework unites OOD robustness with high detection performance (mAP@0.5: 0.947) and enhanced interpretability through Grad-CAM visualizations.
Experimental validation establishes that OOD filtering significantly improves system reliability by preventing false alarms on out-of-distribution inputs while maintaining higher detection accuracy on mammographic data. The present study offers a fundamental foundation for the deployment of reliable AI-based breast cancer detection systems in diverse clinical environments with inherent data heterogeneity.
\end{abstract}


\section{Introduction}

A global health concern, breast cancer is the second-highest cause of cancer related to mortality in women. It has been recorded as the most diagnosed disease in the world in 2020 \cite{intro3a}. According to the World Health Organization, all types of cancer account for 626700 global deaths of women, out of which the breast is the predominant and second leading cause\cite{intro3b}.
If diagnosed in its early development stage, the survival rate are likely to be high and the treatment cost will get reduced \cite{intro1b}. Studies has found that 30\% breast cancer are diagnosed when the size of the mass is 30mm. However if the tumor can identified before it grows more than 10mm then it can be cured completely \cite{Kitchen1998, Gouveia2025}. Due to these facts various campaigns and community health program are conducted where clinical breast examination(CBE) and breast self-examination(BSE) are conducted and taught.Like every cancer, Breast cancer is also divided into two types, Benign and malignant. Benign is non-cancerous but there are cases where these masses turned into the cancer status. Malignant on the other hand tries to spread and infiltrate other tissues.

The main tool for screening the breast cancer is the mammography. The abnormalities in the breast can be captured by passing a low-energy x-rays. The medical images captured from there are assessed by the health experts to interpret the result. According to the current medical practices the examination of the mammograms are done by two radiologist, and by three if the conclusion is hard to draw. Sometimes during the examination the begnin images are misclassified, due to the face that masses may mimic abnormalities of a tumor but not necessarily harmful. The breast cancer classification task is not only challenging due to shape and size of the tumor but also the quality of apparatus and images. The low contrast images might not always show the tumor clearly since its also surrounded by other muscles and blood vessels. 

Due to the rapid advancements in AI, various tools in image analysis, deep learning has been revolutionized. Medical science has also been adapting such tools like computer aided diagnosis (CAD) systems for the classification and detection of various diseases. AI techniques have shown a remarkable aid in the process of diagnosis and treatment using CAD, image interpretation, fusion, registration, segmentation, image-guided surgery, image retrieval and analysis. Such systems provides the practitioners get better understanding of the disease and out body. Many researcher are exploitig the deep-learning models mostly in two fashion: patch-based and ROI-based.In this study we are employing a ROI-based detection technique to identify the regions in the image with a cancerous feature.

In this study, new deep learning models from the YOLO (You Only Look Once) family are employed for tumor detection in mammography images. Different YOLO models were utilized in this study, with YOLOV8 being the primary model because it has a more evolved CNN-based architecture and higher performance in computer vision tasks. YOLOV8 is one of the most stable networks in the YOLO family \cite{intro2}, making it particularly suitable for medical imaging applications. This stability is necessary for maintaining consistency in performance across varied clinical imaging conditions, offering reliable diagnosis regardless of the variation in equipment or image quality. The evaluation between the different versions of YOLO allows for the overall comparison of model performance and robustness in mammographic tumor detection.

The images from the dataset are passed to the model, which places bounding boxes over tumor regions. Additionally, we integrate Out-of-Distribution (OOD) detection to identify anomalous cases that deviate from the training distribution, significantly enhancing the system's reliability by flagging potential misclassifications for closer expert review.
Since diagnosis is a two-way process, without human intervention it is still not considered efficient. The predictions from machines must be scrutinized by experts to assure the identification of disease. Realizing this fact, we are trying to utilize a SOTA detection-based model along with the explainable AI techniques enabling convenient, reliable, and interpretable outcomes. Computers aiding for diagnosis can use the XAI technique to bring the predictions closer to human understanding, hence increasing trust in computers. The combination of YOLO v8's exceptional stability, OOD detection's ability to flag unusual cases, and XAI's interpretability creates a comprehensive framework that addresses the key challenges in automated mammogram analysis.
The primary contributions of this work include:

\begin{enumerate} 
\item Implementation of YOLO architecture optimized for mammographic analysis, achieving superior detection performance.
\item Integration of OOD detection to identify anomalous cases that deviate from the training distribution, enhancing the reliability of the system.
\item Extensive validation on multiple datasets to ensure robustness and clinical applicability.
\item Development of explainable AI mechanisms that provide interpretable visualizations and metrics that align with clinical reasoning.
\item Design of domain adaptation techniques that improve generalization across different mammography equipment and imaging protocols.

\end{enumerate}
By bridging sophisticated AI techniques with practical clinical requirements, this research aims to develop a system that can be meaningfully integrated into breast cancer screening workflows, potentially reducing misdiagnoses and improving early detection rates.

\section{Related Works}
In the last decade, deep learning (DL) methods have transformed medical image analysis, especially in breast cancer screening. Early DL models based on convolutional neural networks (CNNs) made significant gains in both classification accuracy and lesion detection compared to conventional approaches. Recent studies have demonstrated that such models can learn fine imaging cues from full-field mammograms that are typically imperceptible to human interpreters, thereby facilitating earlier and more precise diagnoses \cite{rw1}. 

Early works by LeCun et al. \cite{rw2} introduced the feature extraction ability of convolutional neural networks (CNNs), acting as a precursor for applications ranging from segmentation to classification. DL models have found their widespread application in mammography for the detection of breast cancer, reducing radiologists' workload while improving uniformity in diagnostics. For instance, Gardezi et al. \cite{rw3} demonstrated ML and DL techniques with particular application for breast cancer detection using mammographic data. Similarly, Zheng et al. \cite{rw4} described the progress of the imaging-based AI application in breast cancer screening and diagnosis for clinicians. Esteva et al. \cite{rw5} had proven CNNs' potential in dermatology to open the gates for similar applications in mammography. 
 McKinney et al. \cite{rw6} demonstrated that an AI system capable of surpassing human experts in breast cancer prediction. The authors curated large datasets from both the UK and the USA to evaluate the system's performance. Their results revealed a significant reduction in error rates—with false positives reduced by 5.7\% (USA) and 1.2\% (UK) and false negatives by 9.4\% (USA) and 2.7\% (UK). Moreover, the AI system not only generalized well between the two populations but also achieved an 11.5\% higher AUC-ROC compared to the average radiologist in an independent study of six radiologists, the AI system outperformed all of the human readers.
A sliding window approach was used to scan the whole breast and extract all the possible cancer patches from the image. Several patch-based CNN (VGG16, ResNet50, and InceptionV3) were trained for breast cancer detection, i.e. the classification between positive and negative patches \cite{rw17}.
Despite advancements, problems such as interpretability and domain adaptation between datasets limits the clinical adoption.

The integration of advanced deep learning models, such as the You Only Look Once (YOLO) series, into medical imaging has significantly enhanced the accuracy and efficiency of breast cancer detection in mammograms. Object detection frameworks like Faster R-CNN \cite{rw7} and SSD \cite{rw8} enhanced localization tasks but were slower. The YOLO family solved this by unifying detection into a single network and operating in real time \cite{rw9}.YOLO's real-time object detection feature allows for the detection of regions of interest (ROIs) by bounding boxes, followed by classification, making it a very useful tool for automating diagnostic workflows.


Al-Masni et al. \cite{rw11} proposed a CAD system for mammogram analysis based on YOLO with mammogram pre-processing, convolutional multi-layer feature extraction, mass detection by confidence modelling, and fully connected neural network (FC-NN) classification. Baccouche et al. \cite{rw15} presented YOLO-based strategy for classifying lesions as masses, calcifications, or architectural distortions, supplemented by CycleGAN and Pix2Pix for temporal mammogram change analysis.
Aly et al. \cite{rw16} described the screening mammogram evaluation process as highly monotonous, exhausting, time-consuming, expensive, and highly susceptible to errors for human readers. As a matter of fact, the authors suggested the use of a YoloV3 model for mass detection and classification. With an augmented dataset, they got the most equitable and accurate performance.

Similarly,  Hamed et al.\cite{rw13} integrated YOLOv4 for lesion localization and comparison of feature extraction architectures like ResNet, VGG, and Inception and Su et al.\cite{rw14} proposed two-model fusion of YOLOv5 and LOGO architectures for simultaneous mass detection and segmentation. 

Lan et al. \cite{rw12} proposed an improved YOLOv8 model: YOLOv8-GHOST and YOLOv8-P2 models, tailored for detecting breast mass lesions, achieving a mean Average Precision (mAP) of 71.8\%.


Black-box nature of DL models is problematic in clinical settings.  Post-hoc explanations are provided by XAI methods like LIME \cite{rw20} and SHAP \cite{rw21} by finding salient image regions. Grad-CAM \cite{rw22} and attention mechanisms \cite{rw23} have improved interpretability even more. Explainability for AI models in medical applications is important since it promotes trust and understanding between healthcare professionals. Panwar et al. \cite{rw19} introduced an hybrid model that included Convolutional Neural Networks (CNNs) and Explainable Artificial Intelligence (XAI) techniques for enhancing breast cancer diagnosis using the CBIS-DDSM dataset.The approach not only enhanced the accuracy of diagnosis but also provided interpretable explanations of the predictions of the model. Munshi et al.\cite{rw24} combined U-NET-based image analysis with ensemble models (CNN-RF-SVM) and SHAP explanations for holistic breast cancer diagnosis.

 Prinzi et al. \cite{rw10} implemented a YOLO-based model utilizing transfer learning on public datasets, which effectively detected masses, asymmetries, and distortions in mammographic images. Several YOLO architectures were compared, including YoloV3, Yolov5, and YoloV5-Transformer. In addition, Eigen-CAM was implemented for model introspection and outputs explanation by highlighting all the suspicious regions of interest within the mammogram.

Domain adaptation addresses the challenge of applying models learned from one dataset to other but related datasets, an important challenge to allow generalization of AI applications to different clinical settings.\cite{rw25} proposed the D-MASTER framework, a transformer-based approach for unsupervised domain adaptation in breast cancer detection. This method adaptively masks and reconstructs multi-scale feature maps, enhancing the model's ability to capture reliable features across domains.
Furthermore, Quintana et al. \cite{rw26}, a contrastive learning experiment demonstrated its efficacy as a 2D mammography domain adaptation method with improvement of Area Under the Curve (AUC) from 0.745 to 0.816 on an independent test set. 

\section{Methodology}
\begin{figure}[ht]
\centering
\includegraphics[width=80mm]{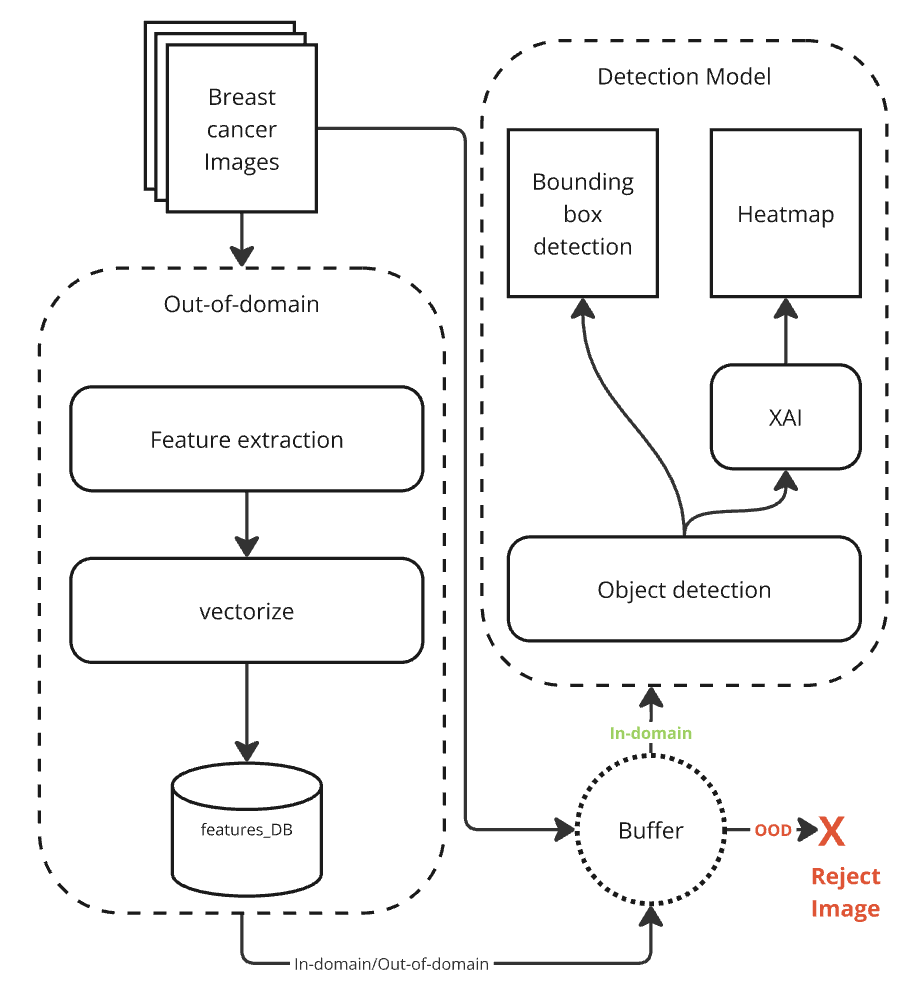}
\caption{Methodology}
\label{fig:methodology}
\end{figure}

The proposed pipeline for breast cancer detection integrates Out-of-Domain (OOD) filtering, YOLOv8-based object detection, and Explainable AI (XAI) to enhance diagnostic accuracy and interpretability in mammographic analysis. The methodology consists of two primary stages: (1) OOD filtering to ensure only relevant mammograms are processed and (2) detection and interpretability enhancement using YOLOv8 and XAI techniques. The overall workflow is depicted in Figure \ref{fig:methodology}.

In the first stage (OOD detection module), the system processes raw medical images, which may include different imaging modalities such as mammograms, ultrasound, CT, and MRI scans. The goal is to ensure that only mammograms are forwarded for analysis, thereby reducing false detections and preventing unnecessary computational overhead. This is achieved through a feature extraction process, where deep learning-based embeddings or handcrafted feature representations are extracted and vectorized. These feature vectors are then compared against a predefined feature database (features\_DB) containing known in-domain and out-of-domain (OOD) examples. Based on this comparison, a buffer module classifies images as either in-domain (mammograms) or OOD (non-mammograms). If an image is classified as OOD, it is immediately rejected, ensuring that only relevant breast cancer images proceed further into the detection pipeline.

In the second stage (Detection and XAI module), the in-domain mammographic images are passed to a YOLOv8-based object detection model for lesion localization. YOLOv8, a state-of-the-art real-time object detection model, predicts bounding boxes around potential cancerous regions with high precision. To enhance the clinical interpretability of AI-driven decisions, we integrate an Explainable AI (XAI) module that generates heatmaps using Grad-CAM. These heatmaps highlight critical regions influencing the model’s decision, allowing radiologists to visually assess the AI’s reasoning. By providing both bounding box annotations and heatmap visualizations, the system ensures greater transparency in AI-based breast cancer detection.

This two-stage methodology is designed to bring AI-based breast cancer detection closer to clinical practice by ensuring that only relevant medical images are processed, improving object detection accuracy, and enhancing interpretability through XAI. The integration of OOD filtering, YOLOv8-based detection, and XAI techniques ensures that the system is robust, interpretable, and clinically viable, making it suitable for deployment in modern radiology workflows.

\subsection*{Out-of-Domain}
This term refers to data outside the original domain or range for which the model was designed or trained. This could include different characteristics or different types of data, not just different distributions. Neural network architectures behave unpredictably when testing on inputs that do not resemble any in their training data. The detection of any OOD inputs is thus of value, in that it might make any overseers aware of limitations in the model's output \cite{meth7}.

For instance, in breast cancer detection systems, OOD scenarios arises when models are applied to unseen patient populations, in various imaging modalities, or pathological conditions that were not sufficiently represented in the training dataset. This may cause degraded performance and possibly unreliable predictions, since a model may not generalize well to novel or unseen data distributions. In medical imaging applications, breast cancer diagnosis, for instance, addressing OOD issues becomes crucial, with large stakes at play in the event of incorrect diagnoses. Our works thus emphasize the development of robust models that can detect such OOD inputs and adapt to them so that the models can work reliably in real-world clinical settings.


\begin{figure}[ht]
\centering
\includegraphics[width=\linewidth]{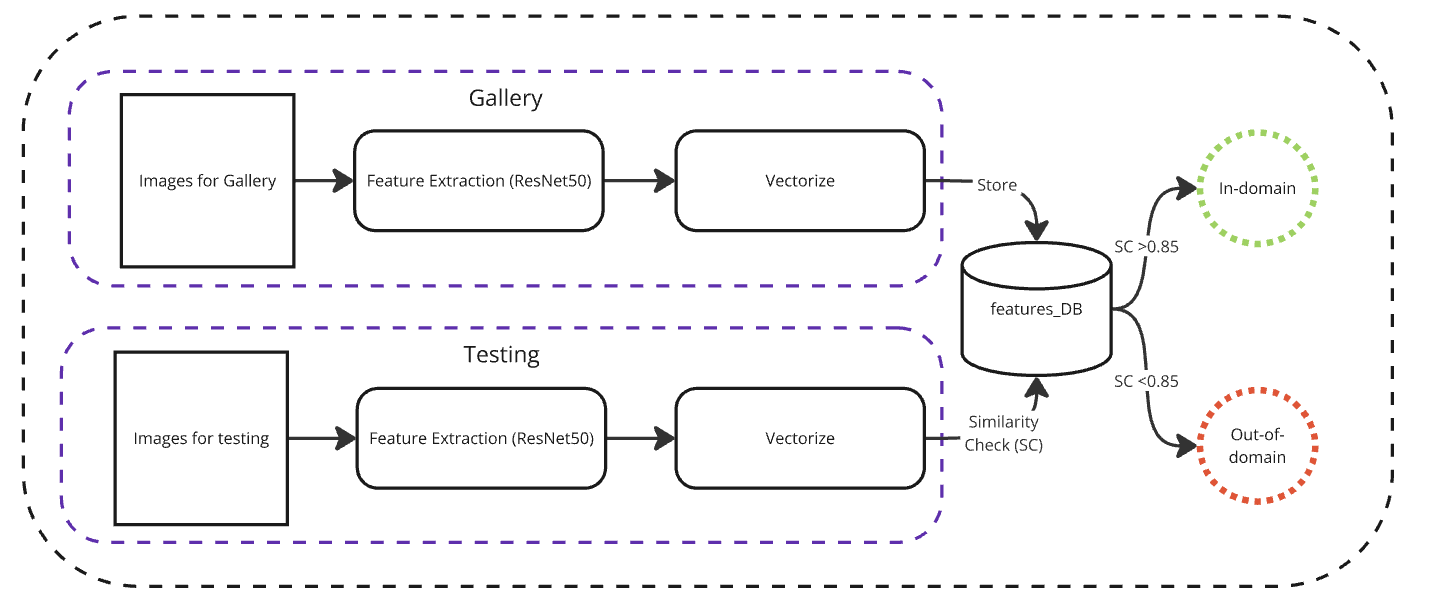}
\caption{Out-of-domain methodology.}
\label{fig:ood}
\end{figure}


The methodology that was followed during OOD detection, involves the following steps:
\begin{enumerate}
    \item \textbf{Gallery Creation}
    \begin{enumerate}
    \item \textbf{In-Domain Gallery Construction:}  ResNet50 is used to analyze a carefully curated dataset of radiological breast cancer images (in-domain). The model extracts high-dimensional feature vectors from the images.

    \item \textbf{Feature Vectorization and Storage:} The features are vectorized and stored within a database, establishing a reference store for similarity comparisons during testing.
    \end{enumerate}   
    \item \textbf{Testing}
    \begin{enumerate}
        \item \textbf{Feature Extraction for Test Images:} During inference, test images are subjected to the same ResNet50-based feature extraction. 

        \item \textbf{Similarity Check mechanism:} Cosine similarity measurement determines the similarity of the test image's feature vector to its nearest neighbors in the database. A predefined threshold value (SC=0.85) determines OOD status:
         \begin{enumerate}
             \item \textbf{In-Domain:} If similarity $\geq 0.85$, the image is labeled as in-domain and then processed by the YOLOv8.

             \item \textbf{Out-of-Domain:}  If similarity $< 0.85$, the image is labeled as OOD, indicating potential anomalies (e.g., artifacts, non-breast tissues, or poor-quality scans).
         \end{enumerate}
    \end{enumerate}
\end{enumerate}

\subsection*{Backbone Architecture Selection}
To select the optimal CNN backbone for OOD detection, we evaluated 12 architectures from five model families: ResNet (ResNet18, ResNet34, ResNet50, ResNet101, ResNet152), VGG (VGG16, VGG19), Inception (InceptionV3), DenseNet (DenseNet121, DenseNet169), and EfficientNet (EfficientNet-B0, EfficientNet-B9). The selection process employed multi-criteria evaluation considering computational complexity, processing efficiency, and detection accuracy.

\subsubsection*{Evaluation Metrics}
Each architecture was assessed using following metrics:

\begin{itemize}
    \item Model Complexity: Parameter count (M) and FLOPs (G)
    \item Computational Efficiency: Feature extraction time and total inference time
    \item Detection Performance: In-domain accuracy and OOD detection accuracy on two test sets
\end{itemize}

\subsubsection*{Statistical Analysis }
We computed mean and median values across all metrics to establish selection benchmarks. The optimal architecture was identified based on: 
\begin{enumerate}
    \item Computational requirements below or near median values,
    \item Detection accuracy above median performance.
    \item Balanced performance across all criteria.
\end{enumerate}
A composite score was calculated using weighted normalization:

\begin{equation}
\begin{aligned}
\text{Composite Score} ={} &   W_1 \times \text{Accuracy} \\
& + W_2 \times \text{Efficiency} \\
& + W_3 \times \text{Robustness}
\end{aligned}
\end{equation}

\subsection*{YOLO}
YOLO families definitely reshaped the landscape in object detection. Moving away from more traditional approaches to two-stage object detection, it first proposed an all-new single-stage scheme that made real-time impressively computation-efficient processing possible. Originally developed by breaking down the input image into grid divisions and thus predicting bounding box and class probabilities, huge saving from computational overhead was possible as against predecessors like Faster R-CNN \cite{rw9}.


YOLO has been pushing the boundaries of object detection technology with each version. Each iteration has incrementally improved model performance, making the algorithms increasingly accessible to developers and researchers \cite{meth1}. The ability of the models to operate on a single GPU and perform efficiently on edge devices democratizes advanced computer vision capabilities.

\subsection*{YOLOv8}
YOLOv8 \cite{meth2}  was another important release from Ultralytics, and it introduced a few critical architectural innovations. Unlike its predecessors, it utilized an anchor-free mechanism of detection, while its backbone network was an innovation borrowed from EfficientNet. This design philosophy prioritized both detection accuracy and computational speed, achieving a delicate balance that differentiated the model from earlier object detection models.


YOLOv8 is similar in structure to previous ones, but brings considerable improvements. More advanced network architectures are included, such as Feature Pyramid Network combined with Path Aggregation Network technologies. In addition, a new and improved annotation interface will make the image labeling process much easier with its new features: auto-labeling, keyboard shortcuts, and configurable hotkey settings.
The FPN component downsamples the spatial dimensions of the input image while expanding its depth along feature channels to build multi-scale feature representations for object detection. PAN architecture supports FPN by the use of skip connections in merging features at varied levels across the network. By working hand in glove with the FPN, this does enable the model to become invariant in identifying an object independent of any scale or form factor, as it is now more capable of handling and concatenating visual information across multiple resolutions \cite{meth1}

\subsection*{YOLOv11}
As a step forward from its predecessors, YOLOv11 has a transformer backbone that identifies long-range dependencies in images and hence improves small object and complex scene detection. The model also has dynamic head design that allows it to adapt to varied complexities of images and employs dual label assignment methods for optimizing speed and performance. They all contribute to YOLOv11 being a handy choice for applications requiring accurate and rapid object detection, e.g., self-driving car navigation and medical image diagnosis \cite{meth3}. 


\subsection*{YOLOv12}
 YOLOv12 introduces an attention-centric architecture that departs from traditional convolutional neural networks. With the implementation of area-based attention mechanisms, YOLOv12 determines and successfully segment feature maps to focus in the important regions within an image. FlashAttention accelerates this process, reducing memory overhead and accomplishing near real-time processing at high resolution. Empirical evaluation confirms that YOLOv12 is more accurate than its predecessors while still rivaling peers with similar speeds. For instance, YOLOv12-N achieves 40.6\% mean Average Precision (mAP) when the inference latency is 1.64 milliseconds on a T4 GPU and outperforms more complex versions like YOLOv10-N and YOLOv11-N by 2.1\% mAP and 1.2\% mAP, respectively, without sacrificing speed \cite{meth4}.


\subsection*{Explainable AI (XAI)}
Explainable AI is an important approach that contributes to the transparency and interpretability of complex machine learning models, especially in high-stakes domains such as healthcare. XAI fills the critical gap between advanced computational techniques and human understanding by providing insight into algorithmic decision-making processes.
XAI methods have become important enablers in the domain of medical imaging for validation and understanding the predictions of artificial intelligence. Methods like LIME, SHAP, CIU and Grad-CAM have enabled researchers and clinicians to visualize and comprehend how deep learning models arrive at a certain diagnosis \cite{meth5}.


Among different visual analytics, Grad-CAM is one of the most powerful XAI techniques applied to image-based analysis. Grad-CAM generates a visual heatmap that highlights the most relevant sections for decision making, helping radiologists investigate and confirm AI predictions together with their underlying justification for diagnostic assessment \cite{meth6}.


XAI techniques are not limited to visualization in medical imaging; these techniques give important insights into model behavior, allowing clinicians to:
\begin{itemize}
    \item Validate the predictions around diagnoses done by AI
    \item Recognize critical features linked to model decisions
    \item Improve confidence in AI systems
    \item Enable more accurate diagnosis facilitated by model interpretability. 
\end{itemize}

Although XAI techniques have brilliant prospects, their clinical utility depends on the generation of valid, relevant, and intuitive explanations that conform to the diagnostic reasoning processes of medical professionals.
Amongst the key milestones that would make AI more transparent, trustworthy, and collaborative in a complex professional environment like healthcare is the continuous development of methodologies related to XAI.

\subsection*{Dataset}

For our study, we utilized mammographic images gathered from the INbreast database. The INbreast database was developed at Porto's Centro Hospitalar de S. João Breast Centre following due ethical approvals from Portugal's National Committee of Data Protection and the Hospital Ethics Committee. The database was specifically designed to make available a means to develop algorithms to detect and diagnose mammographic lesions, addressing the critical need for large digital mammography repositories.The database consists of full-field digital mammograms (FFDM) with thoughtful radiologist marking and is hence extremely rich in terms of research. It comprises 115 cases in total 90 of bilateral breast involvement (providing four images per case) and 25 of mastectomy patients (providing two images per case) \cite{data1}.

For our study, we focused on mass lesions in a total of 1720 images. The images were normalized to 640 × 640 pixels for compatibility with latest YOLO iterations. Auto-orientation was performed during pre-processing without employing any additional augmentation techniques. The dataset was split into training and testing sets in a ratio of 1686:34, respectively.

Moreover, we added an extra OOD dataset to the system, composed of X-ray and MRI images downloaded from Kaggle, representing all the anatomy that was not represented in the training set and including the following: 

\begin{itemize}
    \item MRI images of brain and hands: Providing a variety of anatomic views.
    \item X-ray images: Providing different modality of imaging.
\end{itemize}


In addition to the in-distribution and out-of-distribution data, we used a dataset comprising images representing the various benign states of the breast, including mastitis and fibrocystic changes. We used this to test the ability of our model to rule out invasive breast cancer from other forms of abnormalities.



\subsection*{Performance evaluation metrics}
\subsubsection*{Domain Adaptation metrics}
\begin{itemize}
    \item \textbf{Accuracy:}  Accuracy is a measure of how well a model correctly distinguishes between in-domain  and out-of-domain  samples. The overall accuracy for in-domain (ID) and out-of-domain (OOD) detection is computed as:
\begin{equation}
    \text{Accuracy} = \frac{\text{Total Correctly Classified Images}}{\text{Total Test Images}} 
\end{equation}

\begin{equation}
    = \frac{C_{\text{ID}} + C_{\text{OOD}}}{N_{\text{ID}} + N_{\text{OOD}}} 
\end{equation}

where:
\begin{itemize}
    \item $C_{\text{ID}}$ is the number of correctly classified in-domain images.
    \item $C_{\text{OOD}}$ is the number of correctly classified out-of-domain images.
    \item $N_{\text{ID}}$ is the total number of in-domain test images.
    \item $N_{\text{OOD}}$ is the total number of out-of-domain test images.
\end{itemize}

\end{itemize}

\subsubsection*{Object Detection metrics}
To evaluate the performance of the model, this paper analyzes key validation metrics, incorporating confidence intervals to assess their reliability. The following criteria are considered: Precision-Confidence, Recall-Confidence, Precision-Recall, and F1-Confidence.
Object detection models such as YOLO are evaluated using specific validation metrics that assess their ability to detect and classify objects accurately. The following metrics are used to validate the performance of the model.
\begin{itemize}

    \item {\textbf{Intersection over Union (IoU)}}
Intersection over Union (IoU) measures how well a predicted bounding box overlaps with the ground truth bounding box.

\begin{equation}
    IoU = \frac{\text{Area of Overlap}}{\text{Area of Union}} = \frac{B_{predicted}  \cap B_{groundtruth}}{B_{predicted} \cup B_{groundtruth}}
\end{equation}

A higher IoU indicates better localization accuracy.

    \item {\textbf{Mean Average Precision (mAP)}}
Mean Average Precision (mAP) is a key metric for object detection models. It calculates the area under the Precision-Recall (PR) curve for each class and averages the values.

\begin{equation}
    mAP = \frac{1}{N} \sum_{i=1}^{N} AP_i
\end{equation}

where:
\begin{itemize}
    \item $N$ is the number of object classes.
    \item $AP_i$ is the Average Precision for class $i$.
\end{itemize}

    \item {\textbf{Average Precision (AP)}}
The Average Precision (AP) is computed as the area under the Precision-Recall curve for a specific class.

\begin{equation}
    AP = \int_{0}^{1} P(R) \, dR
\end{equation}

where:
\begin{itemize}
    \item $P(R)$ represents the precision as a function of recall.
\end{itemize}

    \item {Precision-Confidence}
Precision measures the proportion of correctly classified positive instances out of all predicted positive instances. Confidence intervals for precision provide insight into the stability of the model’s precision across different samples.

\begin{equation}
    \text{Precision} = \frac{TP}{TP + FP}
\end{equation}

where:
\begin{itemize}
    \item $TP$ (True Positives): Correctly predicted malignant cases.
    \item $FP$ (False Positives): Benign cases incorrectly predicted as malignant.
\end{itemize}

    \item {\textbf{Recall-Confidence}}
Recall (sensitivity) quantifies the proportion of actual positive cases correctly identified. A recall-confidence interval assesses the reliability of the recall measure.

\begin{equation}
    \text{Recall} = \frac{TP}{TP + FN}
\end{equation}

where:
\begin{itemize}
    \item $FN$ (False Negatives): Malignant cases incorrectly classified as benign.
\end{itemize}

    \item {\textbf{Precision-Recall}}
The Precision-Recall Curve (PRC) is used to assess the model’s performance, especially in imbalanced datasets where the Receiver Operating Characteristic (ROC) curve might not provide a clear picture.

The area under the precision-recall curve (AUC-PR) is computed to quantify the trade-off between precision and recall. Higher AUC-PR values indicate better model performance.

\begin{equation}
    \text{AUC-PR} = \int_{0}^{1} P(R) \, dR
\end{equation}

where $P(R)$ is the precision as a function of recall.

    \item{\textbf{F1-Confidence}}
The F1-score is the harmonic mean of precision and recall, balancing false positives and false negatives. Confidence intervals for the F1-score give a probabilistic range for its stability.

\begin{equation}
    F1 = 2 \times \frac{\text{Precision} \times \text{Recall}}{\text{Precision} + \text{Recall}}
\end{equation}

\end{itemize}
These validation metrics provide a robust assessment of the model’s performance. Confidence intervals offer statistical insight into the model’s stability, ensuring its reliability in medical image classification tasks.

\subsubsection*{Explainable AI (XAI) metrics}
When we can see how models make decisions, we gain insight into their weaknesses, particularly in spotting dataset biases and strengthening defenses against adversarial attacks. By using XAI techniques, specifically Grad-CAM in this research, we help build trust among users like medical professionals regarding the dependability of classification and object detection systems. To evaluate how well XAI model explain decisions, we use several measurement indexes:

\begin{itemize}
    \item \textbf{Matching Ground Truth (MGT) value:} 
    The MGT (Matching Ground Truth) metric evaluates how well a saliency map or heat map aligns with the ground truth mask in identifying important image regions.

    \[
MGT = \frac{n}{p}
\]

In this equation, p represents the total count of pixels marked as important (ones) in the ground truth mask, while n counts how many pixels in the saliency map's brightest regions correctly match these important areas in the ground truth mask. A higher MGT score indicates better performance of the XAI method, as it shows the heat map is accurately highlighting the truly significant regions.

\item \textbf{Pearson Correlation Coefficient (PCC):} The Pearson Correlation Coefficient (PCC) measures the linear relationship between two variables by determining if changes in pixel intensities within highlighted areas of both the ground truth and heat map are correlated, suggesting they emphasize the same regions. PCC values range between -1 and 1, with:
     \begin{itemize}
        \item A value of 1 indicating perfect positive correlation (variables change together in the same direction)
        \item A value of -1 showing perfect negative correlation (variables change in opposite directions)
        \item A value of 0 representing no correlation (no relationship between variables)
    \end{itemize}
 The PCC is calculated using the equation below:
    \begin{equation}
    PCC (u_1,u_2) = \frac{ u_1^T u_2 }{ \| u_1 \| \| u_2 \| }
    \end{equation}
Where \( u_1 \) represents the flattened ground truth mask and \( u_2 \) represents the flattened heat map mask. A high PCC value suggests that the model's interpretation of important image regions closely aligns with the reference ground truth, which is crucial for validating XAI methods. This alignment is especially important in medical imaging applications, where correspondence between the model's focus and ground truth builds confidence in its predictions.

\item \textbf{Root Mean Square Error (RMSE):} RMSE quantifies the average magnitude of errors between two sets of values by calculating the square root of the average squared differences.
The formula for RMSE is:
\begin{equation}
    RMSE = \sqrt{\frac{1}{HW} \sum_{i=1}^{H} \sum_{j=1}^{W} (y_{ij} - y'_{ij})^2 }
\end{equation}
Where, $H$ and $W$ are height and width of an image while $y_{ij}$ is the ground truth mask and $y'_{ij}$ is the heat map mask. A lower RMSE value indicates smaller differences between the ground truth and heat map masks, suggesting the model's interpretation closely aligns with the ground truth and thus implies higher accuracy. Conversely, a higher RMSE value reflects greater discrepancies, indicating that the model's focus is less consistent with the ground truth.
\end{itemize}

\section{Results and Discussion}
\subsection*{Domain Adaptation}
The evaluation results indicate how well the model classifies In-Domain (ID) and Out-of-Domain (OOD) test images. The performance is broken down into three categories:
\begin{itemize}
    \item In-Domain Test Data: These are images that belong to the same domain the model was trained on.
\item Out-of-Domain Test Data (testdata2): Images that belong to a different distribution but are successfully identified as OOD.
\item Out-of-Domain Test Data (testdata3 - other diseases): Another set of OOD images from a different category (e.g., other diseases), also correctly identified.
\end{itemize}
The overall accuracy is calculated based on the total correctly classified images divided by the total test images.




\subsubsection*{Backbone Architecture Selection}
To identify the optimal backbone architecture for our OOD detection module, we evaluated 12 different CNN architectures across multiple performance metrics.
\begin{table}[h!]
\centering
\small
\caption{Model complexity and timing comparison.}
\begin{tabular}{lp{1.3cm}p{1.0cm}p{1.5cm}p{1.2cm}}
\textbf{Model} & \textbf{Parameters (M)} & \textbf{FLOPs (G)} & 
\textbf{Feature Extraction Time (s)} & \textbf{Total Inference Time (s)} \\
\hline
ResNet18       & 11.7 & 1.8  & 175.6  & 199.21 \\
ResNet34       & 21.8 & 3.6  & 283.2  & 354.12 \\
ResNet50       & 25.6 & 4.1  & 335.07 & 335.79 \\
ResNet101      & 44.5 & 7.9  & 581.67 & 606.28 \\
ResNet152      & 60.2 & 11.6 & 826.25 & 859.8  \\
VGG16          & 138.4& 15.5 & 975.27 & 1065.13\\
VGG19          & 143.7& 19.6 & 1254.71& 1208.39\\
Inception v3   & 23.9 & 5.7  & 314.41 & 272.49 \\
DenseNet121    & 8.0  & 2.9  & 277.63 & 240.58 \\
DenseNet169    & 14.1 & 3.4  & 333.6  & 286.28 \\
EfficientNet-b0& 5.3  & 0.39 & 98.35  & 134.07 \\
EfficientNet-b7& 66 & 37 & 692.12 & 823.02 \\
\hline
Mean           & 49.93& 9.46 & 512.08 & 529.58 \\
Median         & 24.75& 4.9  & 334.35 & 344.96 \\
\end{tabular}
\end{table}

\begin{table}[h!]
\centering
\small
\caption{Model performance in terms of accuracy and composite score.}
\begin{tabular}{lp{1.2cm}p{1.2cm}p{1.2cm}p{1.2cm}}
\
\textbf{Model} & \textbf{In-Domain Acc (\%)} & \textbf{OOD-Test1 Acc (\%)} & 
\textbf{OOD-Test2 Acc (\%)} & \textbf{Composite Score} \\
\hline
ResNet18       & 100   & 100    & 81    &  0.77 \\
ResNet34       & 100   & 99.73  & 66.67 &  0.74 \\
ResNet50       & 97.06 & 100    & 100   &  0.79 \\
ResNet101      & 91.18 & 100    & 100   &  0.77 \\
ResNet152      & 88.24 & 100    & 100   &  0.76 \\
VGG16          & 88.24 & 100    & 47.62 &  0.70 \\
VGG19          & 100   & 100    & 52.38 &  0.45 \\
Inception v3   & 100   & 100    & 100   &  0.77 \\
DenseNet121    & 100   & 100    & 85.71 &  0.72 \\
DenseNet169    & 100   & 100    & 95.24 &  0.79 \\
EfficientNet-b0& 76.47 & 100    & 100   &  0.72 \\
EfficientNet-b7& 85.29 & 100    & 80.95 &  0.71 \\
\hline
Mean           & 87.75 & 99.98  & 83.73 &  0.72 \\
Median         & 98.53 & 100    & 88.12 &  0.75 \\
\end{tabular}
\end{table}

\subsubsection*{ResNet50 Selection Justification}

Based on the comprehensive statistical analysis, ResNet50 was selected as the optimal backbone architecture for our OOD detection system. The selection criteria were driven by multiple converging factors:

\begin{enumerate}
    \item Model Complexity: 25.6M parameters (below mean, above median - optimal balance)
    \item  Computational Efficiency: 4.1 GFLOPs (below both mean and median)
    \item Processing Speed: 335.07s feature extraction time (near median performance)
    \item In-Domain Accuracy: 97.06\% (near median of 98.53\%)
    \item OOD Detection: Perfect 100\% accuracy on both OOD test sets
``\item Composite Score: 0.79 (significantly above mean of 0.72 and median of 0.74)
    
\end{enumerate}

\begin{figure}[ht]
\centering
\includegraphics[width=\linewidth]{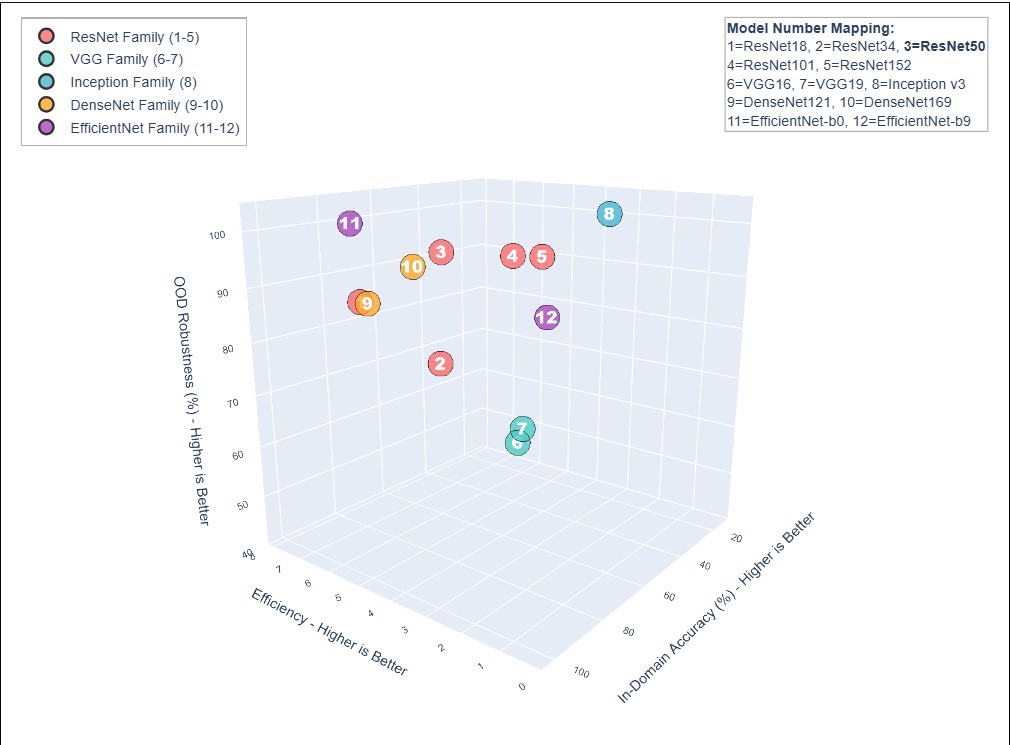}
\caption{Three-dimensional performance visualization of CNN architectures showing the relationship between efficiency, in-domain accuracy, and out-of-distribution detection performance.}
\label{fig:pareto}
\end{figure}

The mean and median analysis strongly supports ResNet50 selection:

\begin{enumerate}
    \item Computational metrics position ResNet50 as efficiently balanced
    \item Accuracy metrics consistently exceed median performance
    \item Composite score ranking places ResNet50 as second-best overall performer
    \item Statistical measures consistently point toward ResNet50 as optimal choice
\end{enumerate}

\begin{figure}[ht]
\centering
\includegraphics[width=\linewidth]{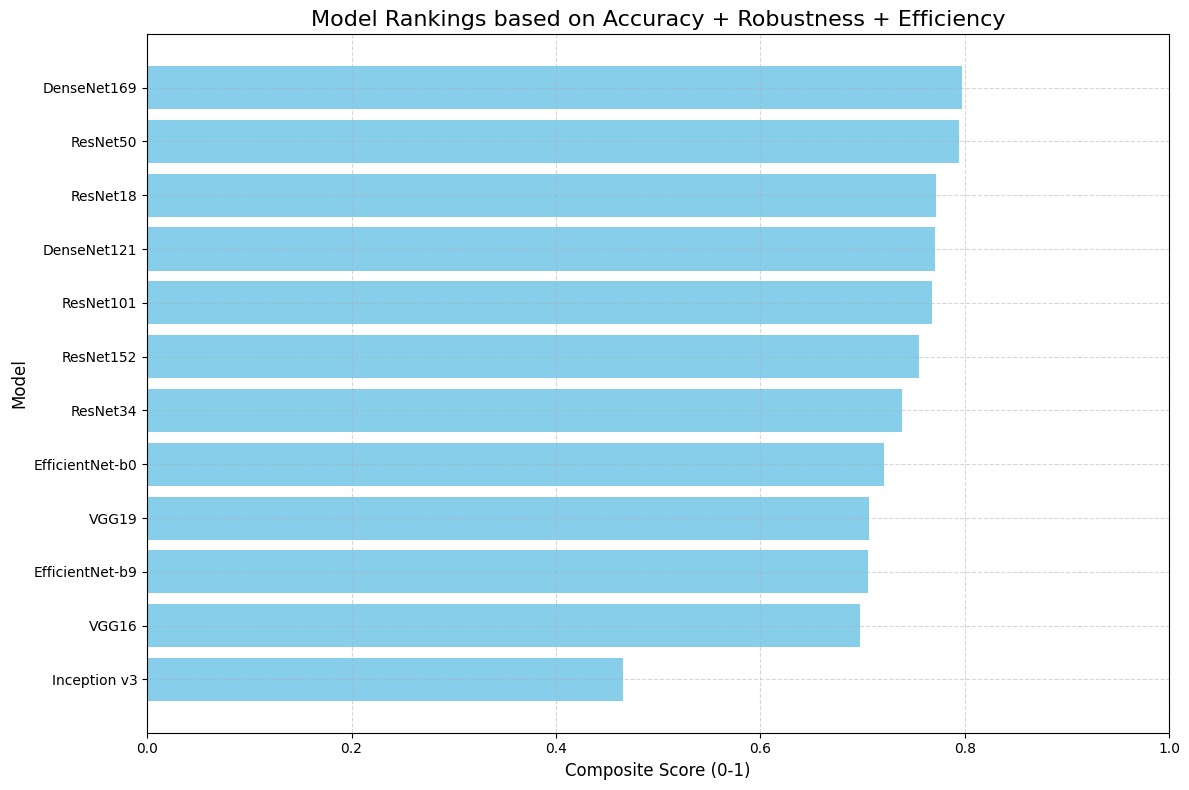}
\caption{Model rankings based on composite score combining accuracy, robustness, and computational efficiency. ResNet50 ranks second overall with a score of 0.794, demonstrating superior balanced performance compared to architectures that excel in individual metrics but lack overall optimization.}
\label{fig:h_chart}
\end{figure}

\subsubsection*{OOD Detection Performance Result}

The  table ~\ref{tab:performance} presents the performance evaluation of the ResNet50 model across In-Domain and Out-of-Domain test datasets:

\begin{table}[h]
    \centering
    \begin{tabular}{p{2.0cm} p{1.0cm} p{1.5cm} p{1.0cm}}
        \textbf{Category} & \textbf{Total Images} & \textbf{Correctly Classified} & \textbf{Accuracy (\%)} \\
        \hline
        In-Domain Test Images & 34 & 33 & 97.06 \\
        Out-of-Domain Test Images (testdata2) & 381 & 381 & 100.00 \\
        Out-of-Domain Test Images (testdata3) & 21 & 21 & 100.00 \\
        \hline
        \textbf{Total Test Images} & \textbf{436} & \textbf{435} & \textbf{99.77} \\
        \hline
    \end{tabular}
    \caption{Model Performance on In-Domain and Out-of-Domain Test Data}
    \label{tab:performance}
\end{table}

\subsubsection*{Observations}
\begin{itemize}
    \item The model performs exceptionally well on \textbf{Out-of-Domain} (OOD) data, achieving \textbf{100\% accuracy} in both OOD test sets.
    \item \textbf{In-Domain accuracy is 97.06\%}, meaning the model correctly classifies most of the in-domain test images.
    \item The \textbf{overall accuracy} is \textbf{99.77\%}, demonstrating high effectiveness in distinguishing between in-domain and OOD images.
\end{itemize}
\subsection*{Detection model}
All the models that came into the line of making our process were trained, validated, and then tested on a single platform: Jupyter Notebook in Google Colab. These platforms allow quick prototyping and deployment of machine learning models on hardware accelerators such as GPUs and TPUs.

\begin{figure}[ht]
\centering
\includegraphics[width=\linewidth]{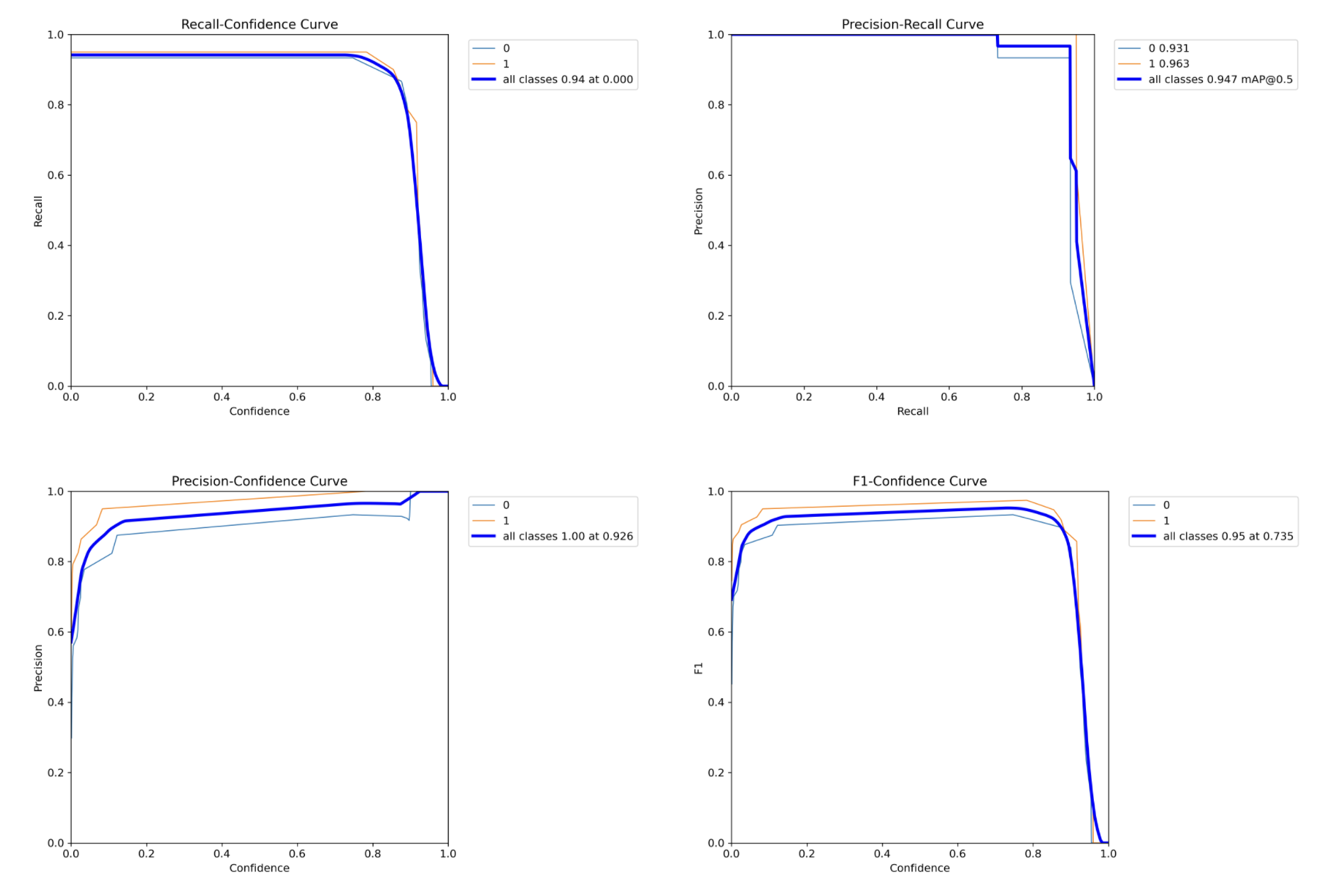}
\label{fig:metrics_1}
\caption{metrics explain more\label{fig:metrics_1}}
\end{figure}

Figure ~\ref{fig:metrics_1} illustrates the normalized confusion matrix for the proposed YOLOv8-based breast cancer detection model. High values along the diagonal (e.g., 0.93 for class 0, 0.95 for class 1) indicate strong classification performance. 

Class 1 performs better than class 0 for most of the confidence thresholds, represented by the orange and light blue curves, respectively. The overall F1 score takes a similar trend, as shown in blue color, reflecting a balanced performance across all classes. The model shows its robustness in maintaining high F1 scores above 0.90 for an inclusive range of thresholds from 0.6 to 0.8. This truly reflects the model's reliability on tasks regarding the detection of breast cancer. 
Above plot illustrates the relationship between model precision and confidence thresholds. As the confidence increases, precision for both classes improves, with an aggregated precision of 1.00 at a confidence threshold of 0.926. This reflects the model's reliability in making high-confidence predictions.
 
The Precision-Recall curve shows the trade-off between precision and recall at different thresholds; class 0 has a precision of 0.931, whereas for class 1, it is 0.963. This gives an mean average precision (mAP) of 0.947 for all classes at a recall threshold of 0.5, reflecting overall strong performance.
 
Above plot shows recall across different levels of confidence: recall stays high across a range of thresholds for confidence but then rapidly drops for very high confidence levels. This reflects balance between sensitivity and confidence in the predictions.
  
\begin{figure}[ht]
\centering
\includegraphics[width=\linewidth]{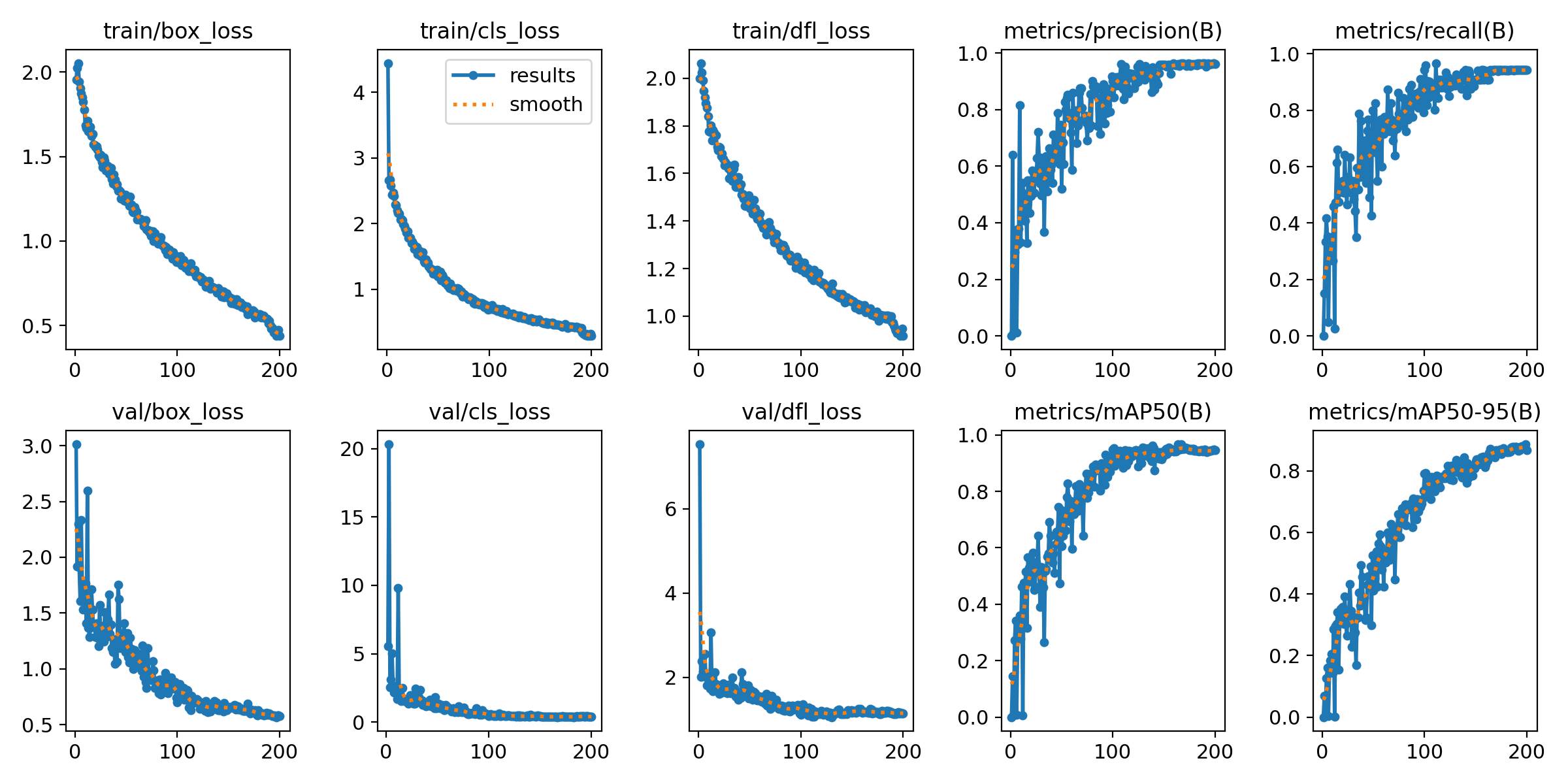}
\label{fig:metrics_2}
\caption{Training and validation performance metrics over 200 epochs, showcasing box loss, classification loss, and DFL loss, alongside precision, recall, and mean average precision (mAP). The results highlight consistent improvement and convergence of the model across both training and validation datasets.\label{fig:metrics_2}}
\end{figure} 

Figure ~\ref{fig:metrics_2} provides a comprehensive visualization of training and validation metrics:
        \begin{itemize}
            \item \textbf{Training Losses:} 
            \begin{itemize}
                \item \textit{Box Loss:} Consistently decreases, indicating improved bounding box regression during training.
                \item \textit{Classification Loss:} Shows a steady reduction, reflecting enhanced classification accuracy.
                \item \textit{Distribution Focal Loss:} Decreases smoothly, ensuring improved localization confidence.
            \end{itemize}
            \item \textbf{Validation Losses:}
            \begin{itemize}
                \item Decreasing trends in box, classification, and distribution focal losses suggest reduced overfitting and improved generalization.
            \end{itemize}
            \item \textbf{Performance Metrics:}
            \begin{itemize}
                \item \textit{Precision:} Steadily increases, highlighting the model's ability to correctly identify positive instances.
                \item \textit{Recall:} Improves consistently, indicating better detection of relevant instances.
                \item \textit{Mean Average Precision (mAP):}
                \begin{itemize}
                    \item \textit{mAP@0.5:} Achieves high values, signifying accurate detection at a 50
                    \item \textit{mAP@0.5:0.95:} Also shows a steady rise, confirming robustness across stricter IoU thresholds.
                \end{itemize}
            \end{itemize}
        \end{itemize}

\subsubsection*{Explainability Grad-CAM}
The first panel shows the original mammogram, while the second panel illustrates the detection results using YOLOv8, highlighting the suspicious lesion with a bounding box and a confidence score of 0.93. The third and fourth panels depict the outputs of Grad-CAM applied to different principal components of the deep learning model's feature space. Principal component analysis (PCA) was used to decompose the model’s learned representations into orthogonal components, capturing variations in the network’s focus areas. The third panel corresponds to the visualization of the third principal component, indicating the most salient features that contribute to the model’s classification, whereas the fourth panel represents the second principal component, which captures a different aspect of the learned features. These heatmaps provide insights into how the model interprets different regions of the mammogram, enhancing interpretability and ensuring that the model’s decisions align with clinical expectations. The integration of PCA with Grad-CAM allows for a more comprehensive understanding of feature importance, facilitating trust and transparency in AI-assisted breast cancer diagnosis.
\begin{figure}[ht]
\centering
\includegraphics[width=\linewidth]{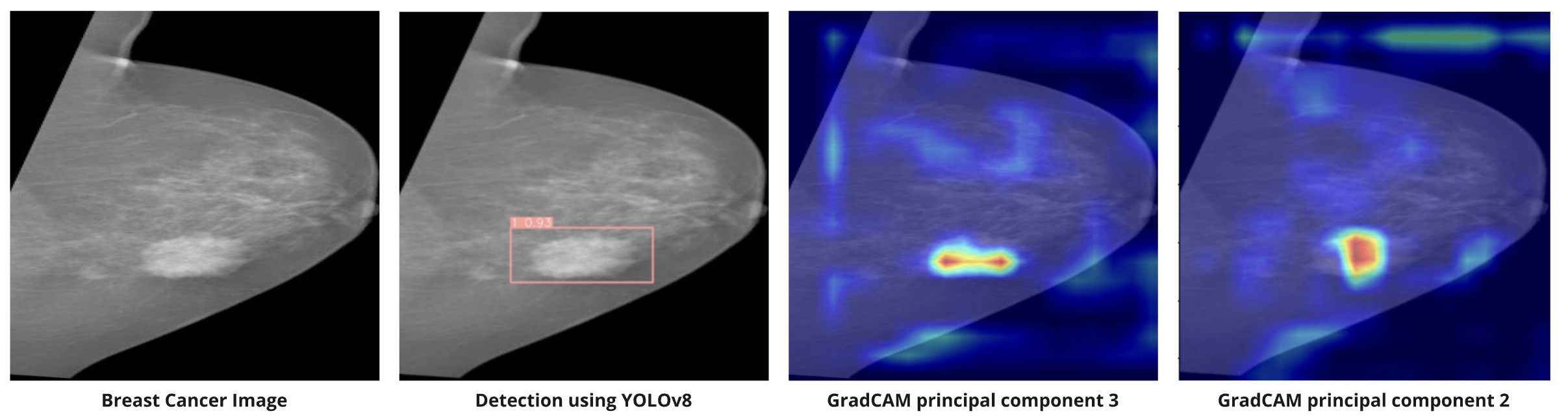}
\label{fig:xai}
\caption{Grad-CAM Heapmap on model's output\label{fig2}}
\end{figure} 

\begin{figure}[ht]
\centering
\includegraphics[width=\linewidth]{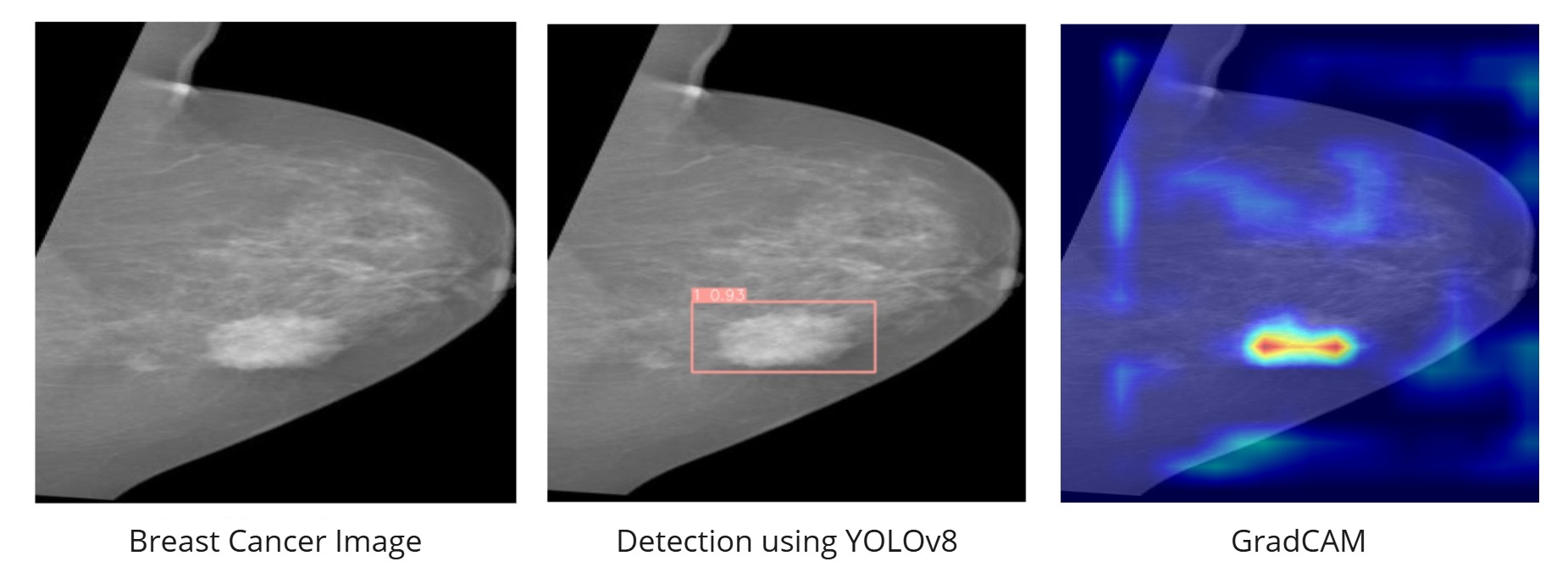}
\label{fig:yolov8}
\caption{Grad-CAM on YOLOv8 output\label{fig:yolov8}}
\end{figure}

\begin{figure}[ht]
\centering
\includegraphics[width=\linewidth]{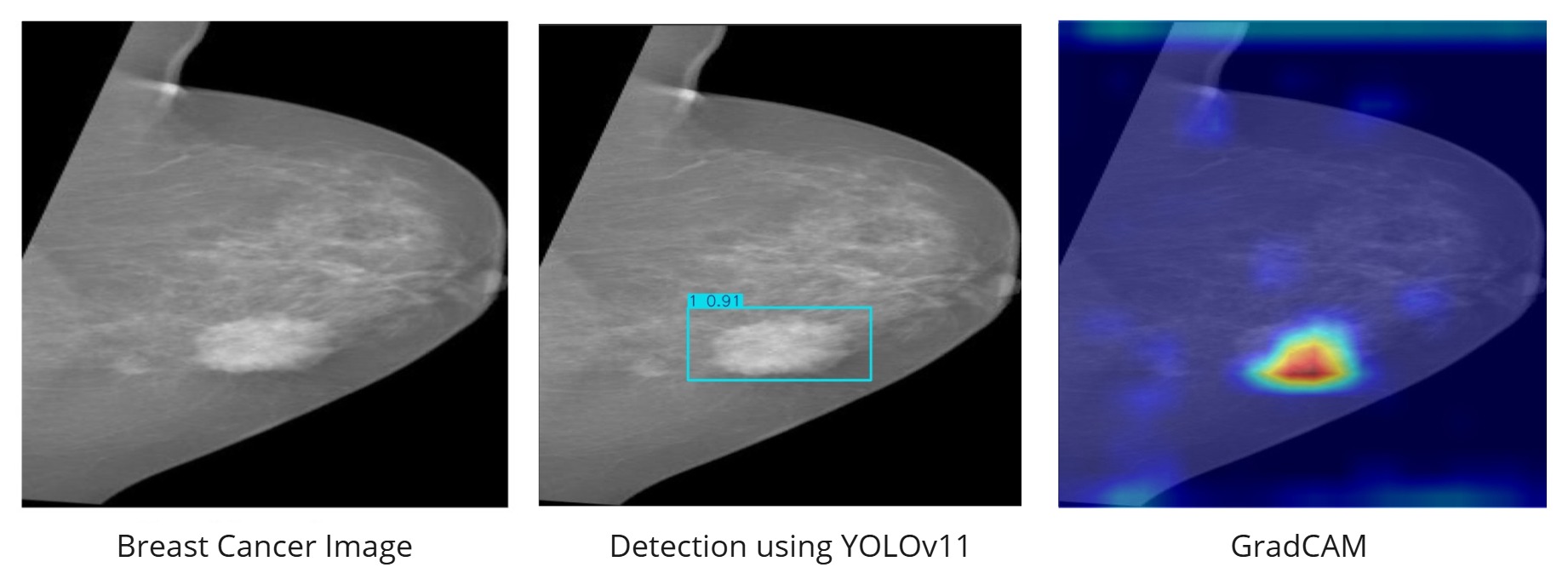}
\label{fig:yolov11}
\caption{Grad-CAM on YOLOv11 output\label{fig:yolov11}}
\end{figure}

\begin{figure}[ht]
\centering
\includegraphics[width=\linewidth]{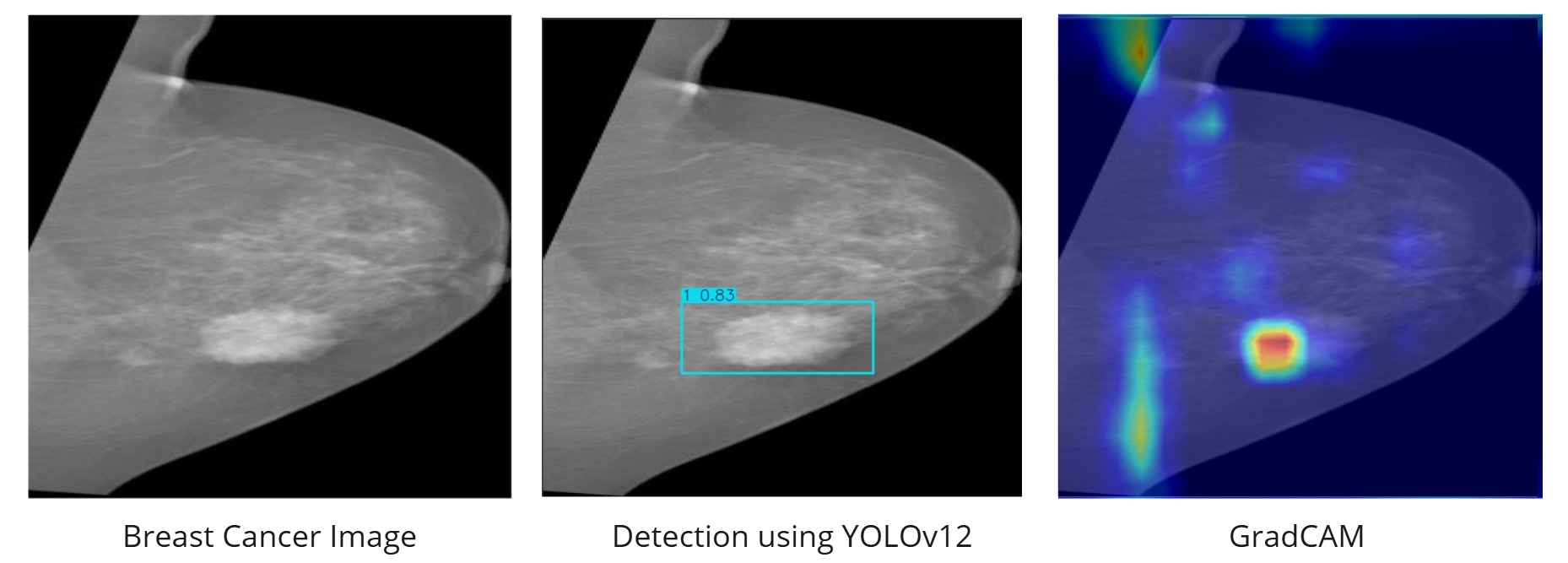}
\label{fig:yolov12}
\caption{Grad-CAM on YOLOv12 output\label{fig:yolov12}}
\end{figure}

\begin{table}[h]
    \centering
    \renewcommand{\arraystretch}{1.5}
    \begin{tabular}{lccc}
        \toprule
        \textbf{Metric} & \textbf{YOLOv8} & \textbf{YOLOv11} & \textbf{YOLOv12} \\
        \midrule
        MGT  & 0.86 & 0.77 & 0.74 \\
        PCC  & 0.31 & 0.39 & 0.31 \\
        RMSE & 0.39 & 0.33 & 0.36 \\
        \bottomrule
    \end{tabular}
    \caption{Evaluation metrics for the Grad-CAM}
    \label{tab:evaluation_metrics}
\end{table}

To analyze the explainability of the given models, Grad-CAM was used to detect regions of interest in mammograms, which are shown in Figures ~\ref{fig:yolov8}, ~\ref{fig:yolov11}, and ~\ref{fig:yolov12} for YOLOv8, YOLOv11, and YOLOv12, respectively. The corresponding evaluation metrics for Grad-CAM visualizations are presented in Table 2. The Matching Ground Truth (MGT) metric is a metric of the precision with which the heatmap aligns with the ground truth mask by quantifying the overlap between salient regions in the saliency map and significant regions in the ground truth mask. The MGT value of greater magnitude indicates that the heatmap heavily distinguishes the actually relevant regions and thus is a critical metric of Explainable AI (XAI) performance.

Among the models, YOLOv8 has the highest MGT score (0.86), which suggests that its heatmap is very close to the ground truth. YOLOv11's MGT score is only slightly lower than that of YOLOv8 (0.77), which indicates a higher but still relevant focus on the lesion, suggesting generalization. YOLOv12, with lowest MGT (0.74), has the worst lesion localization. The Pearson Correlation Coefficient (PCC) for heatmap vs. ground truth lesion area correlation is greatest for YOLOv11 (0.39), indicating its heatmaps are most clinically useful, followed by YOLOv8 and YOLOv12 with lower values at 0.31. In addition, the Root Mean Square Error (RMSE) is lowest for YOLOv11 (0.33), indicating more stable feature localization, while YOLOv8 (0.39) and YOLOv12 (0.36) are more variable.

\section{Conclusion}
Despite the significant advances in deep learning for medical imaging, the fundamental gap remains to address Out-of-Distribution (OOD) detection in breast cancer diagnosis. Current AI-based detection models are prone to depend on the assumption that input images are drawn from the distribution of their training data and are incapable of determining when images differ based on imaging modality variation, patient populations, or acquisition protocols. This monitoring subverts model reliability in real-world clinical application, where data heterogeneity cannot be avoided. Without domain knowledge, deep learning models may generate unreliable choices, leading to false positives, false negatives, and degraded diagnostic errors lowering clinical trust.

To bridge this gap, our approach includes a OOD detection process as a pre-processing step before object detection. Unlike traditional pipelines that execute detection models on various inputs, our approach ensures only domain-specific mammographic images are processed and non-domain-specific or abnormal cases are rejected. We achieved this by using a pre-trained ResNet50 to extract features and created an in-domain gallery of known mammographic patterns. In inference, the test images are filtered for likeness using cosine distance, and if they are under a certain margin, they are marked as OOD and omitted from processing. Pre-filtering not only spares computation but also minimizes misclassifications by avoiding model prediction on non-mammographic or poor inputs.

By enforcing domain knowledge, our system moves AI-based breast cancer detection toward actual medical application. OOD detection enhances model robustness, ensures diagnostic consistency under varying imaging scenarios, and avoids the risk of misleading AI-assisted interpretations. Furthermore, our use of Explainable AI (XAI) techniques, such as Grad-CAM, introduces additional transparency by providing clinicians with visual explanations of the model's predictions.

To foster future research, we have released our framework in PyPi (\url{https://pypi.org/project/out-of-domain-library/}) .

\section{Future Direction}
In future work, the above framework can be extended from breast cancer detection to a wide range of more medical imaging applications. With minor tuning, the same pipeline could assist in detecting lung CT scans for the early pulmonary nodule detection, retinal images for diabetic retinopathy, or dermatological images for skin lesion classification. Besides, the coupling of explainable AI with high-performance detection models offers potential in applications for surveillance of new diseases, where rapid diagnostic support is required during outbreaks. With the integration of other imaging modalities and recurrent updates with different datasets, this platform can become a generalizable computer-aided diagnosis platform, not just beneficial for cancer screening, but also for early diagnosis of new or rare medical conditions.

\bibliographystyle{IEEEtran}  
\bibliography{references}  

\end{document}